\documentclass[11pt]{article}

\usepackage[final]{acl}

\usepackage{times}
\usepackage{latexsym}

\usepackage{booktabs}
\usepackage{multicol}
\usepackage{multirow}

\usepackage{graphicx}
\usepackage[table]{xcolor}
\usepackage[most]{tcolorbox}

\usepackage[T1]{fontenc}

\usepackage{microtype}
\usepackage{amsmath,amssymb}
\usepackage{amsmath}
\usepackage{url}

\definecolor{myred}{RGB}{200,30,30}
\definecolor{myorange}{RGB}{230,140,20}
\definecolor{mygreen}{RGB}{20,140,60}

\usepackage{inconsolata}

\title{Listen, Correct, and Feed Back: Spoken Pedagogical Feedback Generation}

\author{
Junhong Liang,
Yifan Lu,
Ekaterina Kochmar,
Fajri Koto
\\
Mohamed bin Zayed University of Artificial Intelligence, Abu Dhabi, UAE \\
\texttt{\{junhong.liang, yifan.lu, ekaterina.kochmar, fajri.koto\}@mbzuai.ac.ae}
}

\newcommand{\Fhalf}{F$_{0.5}$}

\begin{document}
\maketitle
\begin{abstract}
Grammatical error correction (GEC) and explanation (GEE) have made rapid progress, but real teaching scenarios also require \emph{learner-friendly pedagogical feedback} that is actionable, level-appropriate, and encouraging. We introduce \textbf{SPFG} (\textbf{S}poken \textbf{P}edagogical \textbf{F}eedback \textbf{G}eneration), a dataset built based on the Speak \& Improve Challenge 2025 corpus, pairing fluency-oriented transcriptions with GEC targets and \emph{human-verified} teacher-style feedback, including preferred/rejected feedback pairs for preference learning. We study a transcript-based Spoken Grammatical Error Correction (SGEC) setting and evaluate three instruction-tuned LLMs (Qwen2.5, Llama-3.1, and GLM-4), comparing supervised fine-tuning (SFT) with preference-based alignment (using DPO and KTO) for jointly generating corrections and feedback. Results show that SFT provides the most consistent improvements, while DPO/KTO yield smaller or mixed gains, and that correction quality and feedback quality are weakly coupled. Our implementation is available at \url{https://github.com/Skywalker-Harrison/spfg}.
\end{abstract}

\section{Introduction}
\label{sec:intro}

\begin{figure}[!t]
  \centering
  \includegraphics[width=0.4\textwidth]{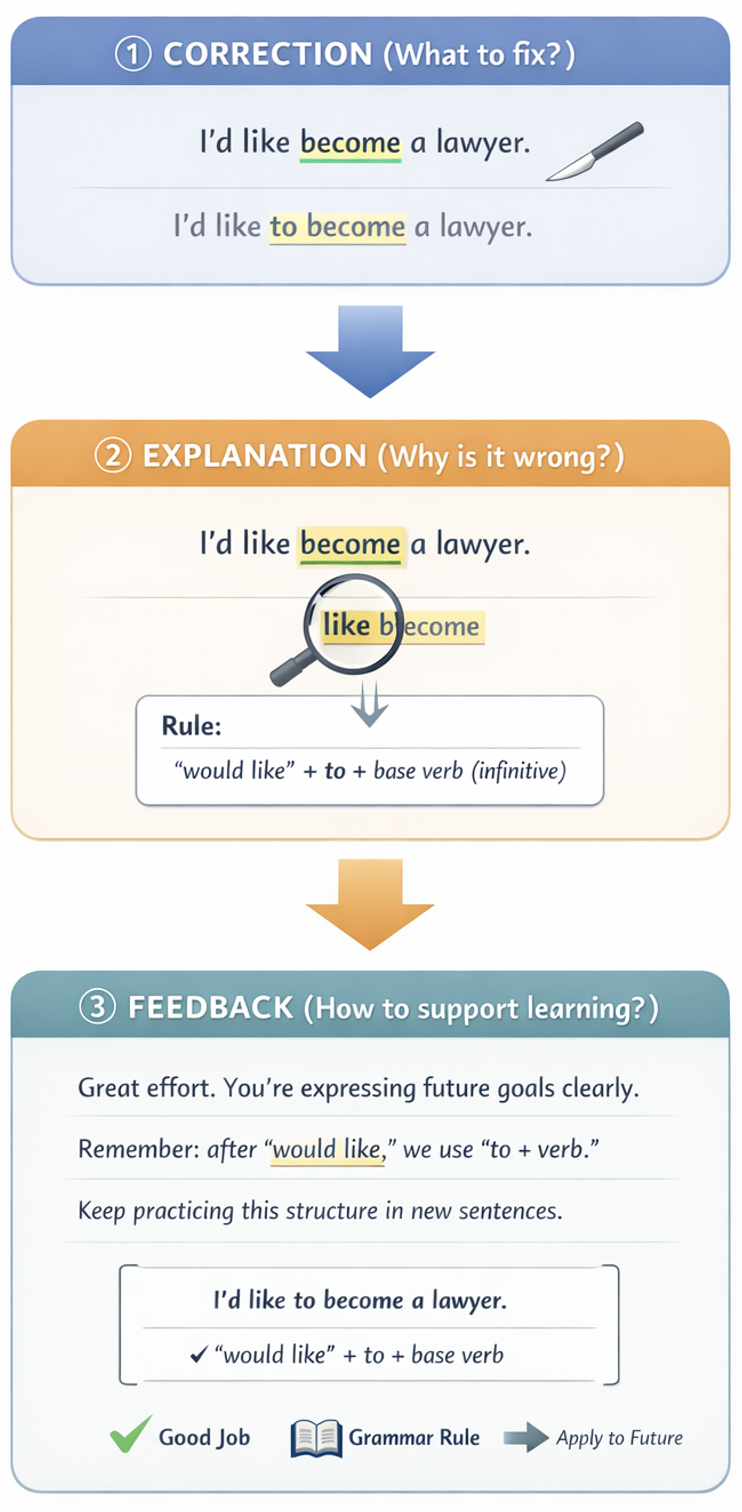}
  \caption{Illustration of three learner-support functions in GEC. \textbf{Correction} rewrites the original utterance into a grammatically well-formed sentence; \textbf{explanation} identifies the error and states the underlying rule; \textbf{feedback} provides learner-friendly guidance with actionable suggestions and encouraging tone.}
  \label{fig:spfg_example}
\end{figure}

Grammatical Error Correction (GEC) has advanced rapidly in recent years, driven by large-scale learner corpora and increasingly powerful sequence-to-sequence and instruction-following models. Beyond correction, Grammatical Error Explanation (GEE) extends GEC systems by not only fixing mistakes but also explaining why a sentence is incorrect in natural language. As GEC systems become widely adopted in educational settings, generating high-quality pedagogical feedback becomes increasingly important. In this work, we investigate \emph{Pedagogical Feedback Generation} (PFG) in the context of Spoken Grammatical Error Correction (SGEC), a setting where speaking plays a central role in second-language (L2) learning, communication, and assessment.

Figure~\ref{fig:spfg_example} illustrates that effective learner support often requires three complementary functions: producing a correction, explaining the underlying issue, and delivering feedback that is actionable, level-appropriate, and encouraging. Pedagogical feedback is critical in user-centered learning systems \cite{williams2024delivering}. 
A good feedback should be faithful to the learner’s errors and the source–target differences, level-appropriate in language and detail, actionable with clear steps the learner can follow, and supportive in tone to encourage improvement \cite{hattie2007power}. However, existing SGEC benchmarks and evaluation protocols focus primarily on correction quality. For instance, the Speak \& Improve Challenge \cite{qian2024speak} evaluates SGEC mainly using edit-based metrics (e.g., MaxMatch/ERRANT-style scoring) \cite{bryant-etal-2017-automatic}, which measure correction fidelity but do not directly assess whether feedback is faithful, clear, or pedagogically helpful.

To address this gap, we study \emph{pedagogical feedback generation} for SGEC and introduce both a dataset and an alignment framework centered on learner-facing quality. We construct \textbf{SPFG} (\textbf{S}poken \textbf{P}edagogical \textbf{F}eedback \textbf{G}eneration) from the Speak \& Improve corpus \cite{qian2024speak}, pairing learner speech with fluency-oriented transcriptions, grammar-corrected targets, and \emph{human-verified} teacher-style feedback. SPFG is designed to support preference learning and includes aligned preferred/rejected feedback pairs. We evaluate three instruction-tuned LLM backbones ({\tt Qwen2.5}, {\tt Llama-3.1}, and {\tt GLM-4}) and compare supervised fine-tuning (SFT) with preference-based alignment methods, including Direct Preference Optimization (DPO) \cite{rafailov2023direct} and Kahneman–Tversky Optimization (KTO) \cite{ethayarajh2024kto}, for jointly generating corrections and pedagogical feedback.

Our main contributions are:
\begin{itemize}
    \item \textbf{Dataset:} We introduce \textbf{SPFG}, a human-verified dataset for SGEC pedagogical feedback generation, including \emph{preferred/rejected} feedback pairs to support preference learning.
    \item \textbf{Alignment:} We provide an empirical study of preference-based alignment for learner-centered feedback, comparing \textbf{DPO} and \textbf{KTO} on top of \textbf{SFT} for jointly generating corrections and pedagogical feedback.
\end{itemize}

\section{Related Works}
\label{sec:related_works}

\subsection{Spoken Grammatical Error Correction}
\label{sec:speech_gec}

The Speak \& Improve Challenge \cite{qian2024speak} provides a shared benchmark for spoken-language learning technologies, spanning automatic speech recognition (ASR), speech language assessment, spoken grammatical error correction (SGEC), and feedback generation. This setting is attractive because it connects correction models to realistic learner data and to downstream educational goals.

To date, many systems submitted to the challenge emphasize upstream recognition and proficiency scoring \cite{banno2025natural,cai25_slate,porwal2025exploring,menevse2025bu}, while comparatively fewer works focus on learner-facing feedback that explains errors and supports revision. In particular, the SGEC task is commonly evaluated with edit-based metrics that quantify correction accuracy at the word or edit level \cite{bryant-etal-2017-automatic}. While such metrics are essential for measuring whether proposed edits are adequate, they do not directly assess pedagogical properties of feedback -- for example, whether the explanation is understandable, level-appropriate, actionable, and encouraging.

\subsection{Pedagogically Aligned Feedback}
\label{sec:pedagogically_aligned_feedback}

A substantial body of work shows that feedback is central to effective learning \cite{wakabayashi2008effect,lu2025belief}. Pedagogically useful feedback is inherently multi-dimensional: it must be not only correct, but also clear, specific, and appropriately calibrated to the learner’s level. Crucially, it should be \emph{explainable} and \emph{faithful} to the learner input—accurately identifying errors and providing grounded rationales rather than unsupported or hallucinated explanations. These requirements make feedback quality difficult to assess automatically and motivate the use of structured evaluation schemes, such as rubric-based or preference-based supervision \cite{scarlatos2024improving}, as well as learning-science-informed principles for effective instruction.

With the rise of large language models (LLMs), there has been increasing interest in generating feedback at scale, alongside concerns about maintaining pedagogical rigor and alignment. One line of work introduces \emph{structure and guardrails} to constrain LLM behavior: for example, MWPTutor \cite{pal2024autotutor} incorporates explicit tutoring constraints, while CLASS \cite{sonkar2023class} formalizes learning-science-inspired design principles. Another line of work focuses on improving feedback quality through optimization and system design. \citet{scarlatos2024improving} propose rubric-based evaluation combined with preference optimization to enhance correctness and pedagogical alignment, while \citet{peng-etal-2025-kele} demonstrate the effectiveness of structured teaching strategies via multi-agent Socratic tutoring. More broadly, multi-agent critique frameworks \cite{jordan2025magic} and rationale-based approaches \cite{cao2025rationalize} aim to produce more interpretable and pedagogically grounded feedback.

These challenges are particularly pronounced in language learning tasks, where feedback must closely track learner errors. In grammatical error correction (GEC) and its extension to spoken GEC (SGEC), feedback must not only propose corrections but also provide accurate, fine-grained explanations aligned with the original input. Recent work explores explanation-aware methods for multilingual GEC \cite{li2025explanation}, revisits the role of LLMs in Chinese GEC \cite{li2025rethinking}, and introduces benchmarks for edit-level explainability \cite{ye2025excgec}. Related studies extend feedback beyond sentence-level correction \cite{vainikko2025paragraph} and to low-resource settings such as Indonesian \cite{10.1007/978-981-95-3343-5_43}. Despite these advances, ensuring that feedback is both pedagogically effective and faithfully grounded in learner input remains a central challenge, particularly in SGEC where spoken input introduces additional variability.

\section{SPFG Data Construction}
\label{sec:spfg_data_construction}

\begin{figure}[t]
  \centering
  \includegraphics[width=0.48\textwidth]{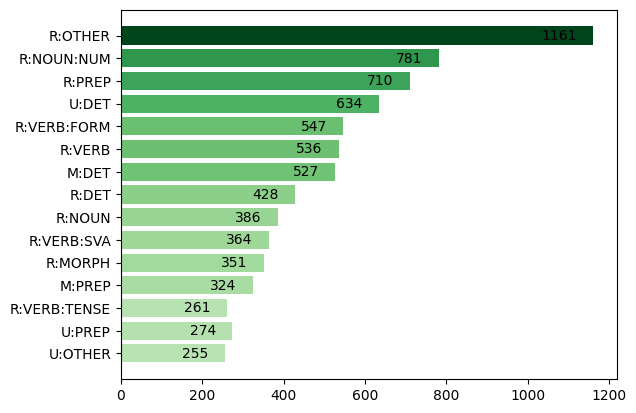}
  \caption{Distribution of the top 15 grammatical error types in the dataset. The color intensity corresponds to the frequency of each error type, with darker shades indicating higher counts.}
  \label{fig:tpo_15_error_types}
\end{figure}

\begin{table}[t]
\centering
\scalebox{0.8}{%
\begin{tabular}{lrrrr}
\toprule
\textbf{Split} & \textbf{\# Audio} & \textbf{Time (hrs)} & \textbf{\# Words} & \textbf{Avg. Words} \\
\midrule
Train & 4,285 & 30.24 & 151,859 & 35.44 \\
Dev   & 500   & 3.53  & 17,739  & 35.48 \\
Eval  & 2,793 & 20.40 & 102,108 & 36.56 \\
\bottomrule
\end{tabular}
}
\caption{Summary statistics of the dataset across different splits.}
\label{tab:data_stats}
\end{table}

\begin{figure*}[t]
  \centering
  \includegraphics[width=\textwidth]{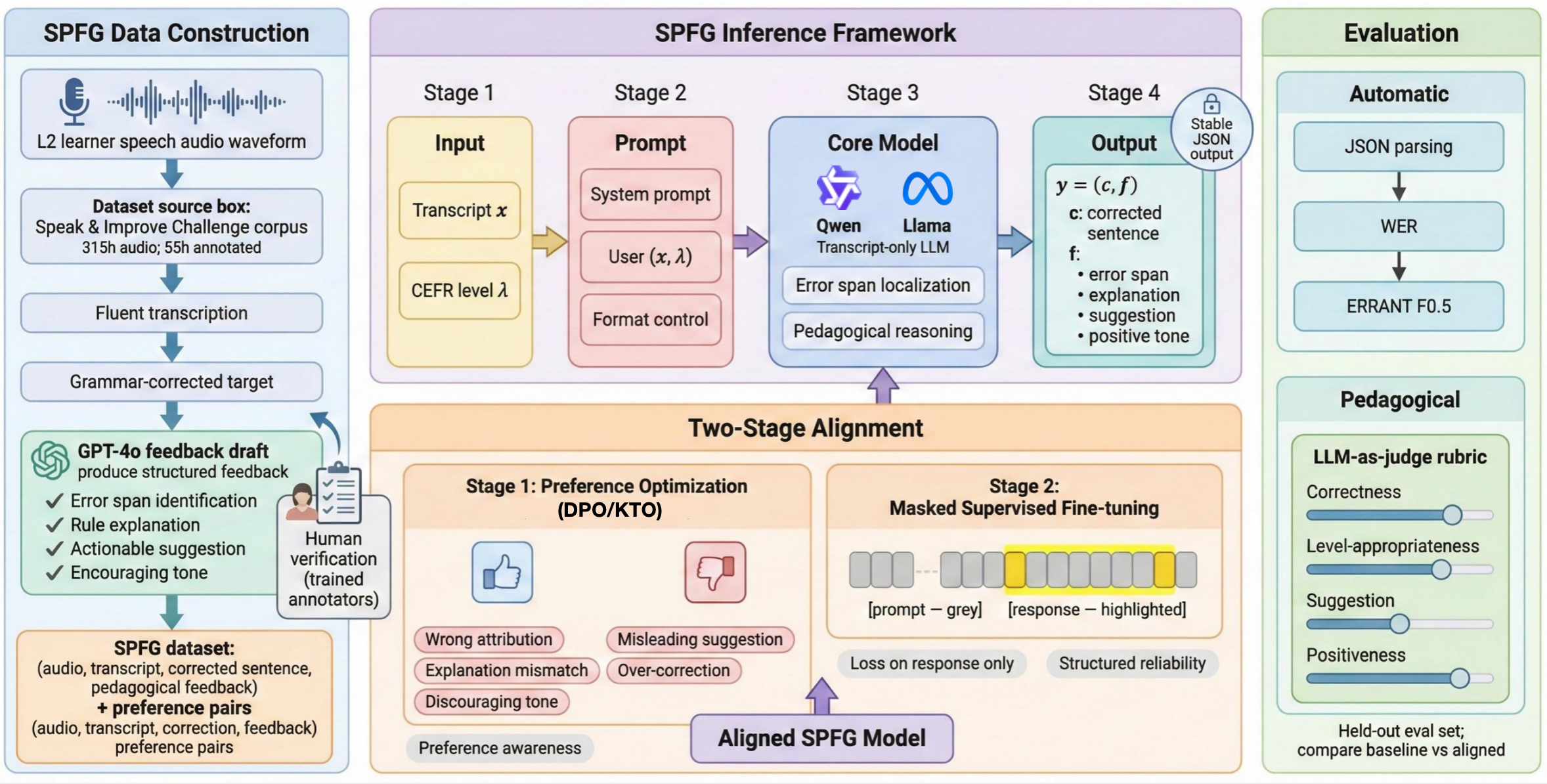}
  \caption{Overview of SPFG. {\bf Left}: data construction from Speak \& Improve (S\&I), producing GEC targets and human-verified pedagogical feedback (see Section~\ref{sec:spfg_data_construction}). {\bf Middle}: inference format and a two-stage training pipeline that combines preference-based alignment and supervised fine-tuning (SFT) to jointly generate corrections and feedback (see Section~\ref{sec:methodology}). {\bf Right}: evaluation with automatic metrics (e.g., WER and ERRANT) and pedagogical quality assessment using an LLM-as-a-judge (see Sections~\ref{sec:experiments}).}
  \label{fig:spfg_overview}
\end{figure*}

We construct SPFG (\textbf{S}poken \textbf{P}edagogical \textbf{F}eedback \textbf{G}eneration) from the Speak \& Improve Challenge corpus \cite{qian2024speak}, a large-scale dataset of L2 English learner speech designed to support speech assessment and feedback research. The corpus contains 315 hours of learner audio recorded at 16~kHz, collected from real users of the Speak \& Improve platform between 2019 and 2024. A 55-hour subset is carefully transcribed and annotated, including human-provided grammatical error corrections.

\paragraph{Example selection.}
In SPFG, we focus on utterances that have both a \emph{fluency-oriented transcription} (\textit{source}) and a corresponding \emph{grammar-corrected sentence} (\textit{target}), which directly supports correction and feedback generation.

\paragraph{Pedagogical feedback generation and verification.}
To obtain learner-facing pedagogical feedback, we employ {\tt GPT-4o} \cite{hurst2024gpt} to generate teacher-style responses conditioned on each paired (\textit{source}, \textit{target}) instance. We utilize this model for its ability to adhere to complex, multi-step instructions while maintaining the supportive, natural tone essential for educational settings. Specifically, the model is prompted to provide: (i) explicit identification of error spans, aligning with Grammatical Error Detection (GED); (ii) actionable corrections, as in Grammatical Error Correction (GEC) \cite{ye2025excgec}; (iii) explanations of the underlying linguistic rules, consistent with Grammatical Error Explanation (GEE) \cite{song2024gee}; and (iv) pedagogical suggestions. By synthesizing these four components, the model produces feedback that is both corrective and pedagogically restorative. The prompts can be found in Appendix \ref{sec:appendix:prompt}. 

The feedback is then \emph{verified by a trained human annotator} who holds a bachelor’s degree in English and a master’s degree in Language Education, with extensive experience in Teaching English as a Second Language (TESOL). To ensure factual correctness and pedagogical usefulness (e.g., correcting wrong attributions, removing misleading advice, and avoiding unnecessary over-corrections). We randomly sample 100 GPT-4o feedback items and ask the expert annotator to judge whether each item is factually correct and pedagogically appropriate. The annotator accepts 91\% of sampled items as satisfactory. This suggests the high quality of the generated feedback. The readability analysis is available in Appendix \ref{apdx:readability_analysis}.

Following the official evaluation protocol, we keep the evaluation set unchanged and adjust the train/dev split to obtain a more balanced development set for model selection. Data splits are described in Table~\ref{tab:data_stats}. 

\paragraph{Error type distribution.} Next, we analyze error type distribution following the definition of \citet{bryant-etal-2017-automatic}. A detailed description and examples can be found in Appendix \ref{apdx:errant_error_code_types}. The results related to the top 15 error types are shown in Figure \ref{fig:tpo_15_error_types}. It shows that \texttt{R:OTHER} (1161), \texttt{R:NOUN:NUM} (781), and \texttt{R:PREP} (710) are the three most frequent error types. In addition, errors related to determiners (\texttt{DET}) and prepositions (\texttt{PREP}) -- including missing (\texttt{M}), replacement (\texttt{R}), and unnecessary (\texttt{U}) forms -- consistently appear with high frequencies across the dataset. This pattern suggests that article usage and preposition selection remain major challenges in spoken English, especially for non-native speakers. The high occurrence of these error categories reflects the inherent difficulty of mastering function words, which are often less semantically salient but highly context-dependent. The error type distribution across different CEFR levels is available in Appendix \ref{apdx:error_type_across_cefr_groups}.

In Table~\ref{tab:cefr_feedback_stats}, we report generated feedback statistics across different Common European Framework of Reference for Languages (CEFR) levels \cite{council2001common}. These levels provide a standardized basis for describing language proficiency and are essential for tailoring pedagogical feedback to a learner's specific competency stage.

\paragraph{Negative Sample Generation for Preference Training} To enable preference-based alignment (e.g., DPO/KTO), we construct preference pairs consisting of a \emph{chosen} response (human-verified feedback) and a \emph{rejected} response (synthetically degraded feedback) under the same input context. The rejected feedback is designed to remain fluent and relevant, while being systematically inferior along targeted pedagogical dimensions listed below, yielding a clear and controllable preference signal. The prompt is available in Appendix \ref{sec:appendix:dpo_prompts}.

We adopt a roughly balanced distribution to ensure that no single pedagogical defect dominates the preference signal. The corruption types include:
(1) {\em Wrong Attribution} (25.1\%), where the correction is plausible but the grammatical category or rule is incorrectly identified;
(2) {\em Explanation Mismatch} (25.3\%), where the explanation appears reasonable in isolation but does not logically justify the provided correction;
(3) {\em Discouraging Tone} (14.2\%), where the feedback is demotivating or judgmental despite being grammatically correct;
(4) {\em Misleading Suggestions} (9.5\%), where the advice sounds helpful but promotes incorrect usage; and
(5) {\em Over-correction} (25.9\%), where unnecessary edits are applied to already acceptable learner utterances.

The proportions are motivated by a preliminary qualitative inspection of zero-shot \texttt{Qwen2.5-7B-Instruct} outputs. This analysis indicates that incorrect attribution and over-correction are the most common failure modes, while tone- and suggestion-related issues occur less frequently. Accordingly, we assign higher proportions to these prevalent structural errors, while still maintaining coverage of secondary pedagogical dimensions.

Using the prompt described in Appendix \ref{sec:appendix:dpo_prompts}, we generate 4,285 negative samples. From these, we randomly select 100 samples for human evaluation. The annotator agrees with 87\% of the generated negative samples, suggesting that our generation procedure produces largely valid and reliable negative instances. 


\begin{table}[t]
\centering
\small
\begin{tabular}{c c c}
\toprule
\textbf{CEFR} & \textbf{Percentage} & \textbf{Avg.\ Len} \\
\midrule
A2     & 1.0\% / 1.8\% / 1.2\% & 135.5 / 152.2 / 166.8 \\
A2--B1 & 3.0\% / 3.5\% / 2.6\% & 154.7 / 166.9 / 142.4 \\
B1     & 10.9\% / 10.5\% / 8.4\% & 159.2 / 180.1 / 165.2 \\
B1--B2 & 20.7\% / 22.7\% / 23.8\% & 166.0 / 182.8 / 168.4 \\
B2     & 24.7\% / 26.2\% / 23.4\% & 170.1 / 186.3 / 175.7 \\
B2--C1 & 25.1\% / 23.2\% / 28.8\% & 166.9 / 186.6 / 162.8 \\
C1     & 11.9\% / 9.5\% / 9.4\% & 170.1 / 189.1 / 160.4 \\
C1--C2 & 2.8\% / 2.6\% / 2.4\% & 178.1 / 175.0 / 151.9 \\
\bottomrule
\end{tabular}
\caption{Distribution of CEFR levels and average feedback length (in words) across train (4,285), eval (2,793), and dev (500) splits. CEFR levels range from A2 to C1--C2.}
\label{tab:cefr_feedback_stats}
\end{table}

\begin{table*}[!t]
  \centering
  \small
  \setlength{\tabcolsep}{6pt}
  \renewcommand{\arraystretch}{1}

  \definecolor{HeaderBg}{RGB}{235,240,248}
  \definecolor{GroupBg}{RGB}{245,245,245}

  \begin{tabular}{l c ccc ccccc}
  \toprule
  Model
  & WER $\downarrow$
  & \multicolumn{3}{c}{ERRANT $\uparrow$}
  & \multicolumn{5}{c}{Feedback Quality (LLM-judge) $\uparrow$} \\
  \cmidrule(lr){3-5}
  \cmidrule(lr){6-10}
  \rowcolor{HeaderBg}
  & & $P$ & $R$ & $F_{0.5}$
  & Correct. & Level & Sugg. & Posit. & Avg. \\
  \midrule

  \rowcolor{GroupBg}
  \multicolumn{10}{c}{\textbf{Proprietary Models}} \\
  DeepSeek-Chat & 21.74 & 43.8 & 58.6 & 46.1 & 4.97 & 4.99 & 4.86 & 4.99 & 4.95 \\
  Gemini-2.5-Flash   & 25.34 & 40.0 & 54.7 & 42.3 & 5.00 & 4.98 & 4.85 & 5.00 & 4.96 \\
  Qwen-Plus     & 38.76 & 25.7 & 44.9 & 28.1 & 4.97 & 4.94 & 4.90 & 4.79 & 4.90 \\

  \midrule
  \rowcolor{GroupBg}
  \multicolumn{10}{c}{\textbf{Open-source LLMs}} \\

  Qwen-2.5                     & 17.97 & 50.9 & 55.3 & 51.7 & 4.32 & 4.69 & 4.29 & 4.87 & 4.55 \\
  Qwen-2.5 (SFT)               & 13.54 & 61.0 & 60.0 & 60.8 & \textbf{4.65} & 4.92 & \textbf{4.54} & \textbf{5.00} & 4.78 \\
  Qwen-2.5 (DPO+SFT)           & 12.76 & 63.9 & 59.6 & 63.0 & \underline{4.60} & 4.93 & \underline{4.47} & \textbf{5.00} & \underline{4.75} \\
  Qwen-2.5 (KTO+SFT)           & 14.45 & 57.9 & 59.5 & 58.2 & 4.59 & 4.86 & \textbf{4.54} & \textbf{5.00} & \underline{4.75} \\
  \addlinespace

  Llama-3.1                    & 33.69 & 30.8 & 46.3 & 33.1 & 4.18 & 4.31 & 4.02 & 4.21 & 4.18 \\
  Llama-3.1 (SFT)              & \textbf{11.24} & 68.3 & \textbf{63.9} & \textbf{67.4} & 4.59 & \underline{4.94} & 4.38 & \textbf{5.00} & 4.73 \\
  Llama-3.1 (DPO+SFT)          & 12.14 & 65.7 & 61.3 & 64.8 & 4.47 & 4.85 & 4.35 & \textbf{5.00} & 4.67 \\
  Llama-3.1 (KTO+SFT)          & 12.19 & 65.0 & 61.8 & 64.3 & 4.50 & 4.82 & 4.41 & \textbf{5.00} & 4.68 \\
  \addlinespace

  GLM-4                    & 26.98 & 36.9 & 50.0 & 38.9 & 4.53 & 4.78 & 4.24 & 4.85 & 4.60 \\
  GLM-4 (SFT)              & \underline{11.28} & 67.9 & \textbf{64.4} & \underline{67.1} & \textbf{4.65} & \textbf{4.96} & \textbf{4.54} & \textbf{5.00} & \textbf{4.79} \\
  GLM-4 (DPO+SFT)          & 11.86 & \textbf{68.6} & 58.7 & 66.3 & 4.57 & 4.87 & 4.39 & \textbf{5.00} & 4.71 \\
  GLM-4 (KTO+SFT)          & 11.49 & \underline{68.5} & 61.0 & 66.9 & 4.57 & 4.89 & 4.36 & \textbf{5.00} & 4.70 \\

  \bottomrule
  \end{tabular}
  \caption{Grammar correction and feedback generation results across different models. WER controls ASR quality; ERRANT evaluates grammatical
  correction accuracy; feedback quality is assessed by an LLM judge along four pedagogically motivated dimensions, with Avg.
   denoting their mean. The best results for Open-source LLMs are marked in 
   \textbf{bold} and the second best results are \underline{underlined}.}
  \label{tab:agee-feedback}
\end{table*}

\section{Methodology: Two-Stage Training with Token Masking}
\label{sec:methodology}

\subsection{Problem Formulation}

We address the task of SGEC with pedagogical feedback generation for English language learners. Given a source utterance $x \in \mathcal{X}$ at proficiency level $\lambda \in \{\textnormal{A1}, \ldots, \textnormal{C2}\}$, our objective is to learn a model $f_\theta: \mathcal{X} \times \lambda \to \mathcal{Y}$ that produces structured output $y = (c, f)$, where $c$ denotes the grammatically corrected sentence and $f$ represents constructive pedagogical feedback.

\subsection{Two-Stage Training Framework}

Our approach adopts a two-stage training paradigm that integrates preference-based optimization with supervised fine-tuning.

\paragraph{Stage 1: Preference Optimization (DPO/KTO).}

We initialize a pre-trained language model $\mathcal{M}_0$ (\texttt{Qwen2.5-7B-Instruct}) and apply parameter-efficient fine-tuning via Low-Rank Adaptation (LoRA)~\citep{hu2021lora} with rank $r$ and scaling factor $\alpha$. 

Given a preference dataset $\mathcal{D}_{\textnormal{pref}} = \{(x_i, y_i^w, y_i^l)\}_{i=1}^{N_{\textnormal{pref}}}$, where $y_i^w$ and $y_i^l$ denote preferred and dispreferred outputs, respectively, we perform preference optimization. In the case of Direct Preference Optimization (DPO)~\citep{rafailov2023direct}, we define:

\begin{equation}
\Delta_\theta(x,y^w,y^l)
=
\log \frac{\pi_\theta(y^w\mid x)}{\pi_{\textnormal{ref}}(y^w\mid x)}
-
\log \frac{\pi_\theta(y^l\mid x)}{\pi_{\textnormal{ref}}(y^l\mid x)}
\end{equation}

\begin{equation}
\mathcal{L}_{\textnormal{pref}}(\theta) 
= 
-\mathbb{E}_{(x,y^w,y^l)\sim\mathcal{D}_{\textnormal{pref}}}
\bigl[\log \sigma(\beta \Delta_\theta(x,y^w,y^l))\bigr]
\end{equation}

\noindent where $\pi_\theta$ denotes the policy model, $\pi_{\textnormal{ref}}$ is the frozen reference model $\mathcal{M}_0$, $\beta$ is a temperature parameter, and $\sigma(\cdot)$ is the sigmoid function.

This objective can be replaced by alternative preference optimization methods such as KTO, which similarly leverage preference signals to align model outputs with desired behaviors. In both cases, the goal is to improve the model's ability to rank high-quality corrections and feedback over inferior ones.

\paragraph{Stage 2: Supervised Fine-Tuning with Token Masking.}

After preference optimization, we merge the learned LoRA adapter $\mathcal{A}_{\textnormal{pref}}$ into the base model and initialize a new LoRA adapter. We then perform supervised fine-tuning on a labeled dataset of $(x, y)$ pairs.

\textbf{Token Masking Strategy.} For each training instance $(x, y)$, we construct a conversational sequence:
\begin{equation}
\mathbf{s} = \text{concat}(\mathbf{p}_{\textnormal{sys}}, \mathbf{p}_{\textnormal{usr}}(x, \lambda), \mathbf{r}_{\text{asst}}(y))
\end{equation}

\noindent where $\mathbf{p}_{\textnormal{sys}}$ is the system prompt, $\mathbf{p}_{\textnormal{usr}}(x, \lambda)$ encodes the user query with source text $x$ and proficiency level $\lambda$, and $\mathbf{r}_{\text{asst}}(y)$ represents the assistant response in structured JSON format.

After tokenization, let $\mathbf{s} = [s_1, s_2, \ldots, s_T] \in \mathcal{V}^T$, where $\mathcal{V}$ is the vocabulary and $T$ is the sequence length. We define a binary mask $\mathbf{m} \in \{0,1\}^T$ as:

\begin{equation}
m_t =
\begin{cases}
0 & \text{if } 1 \leq t \leq T_p \quad \text{(prompt region)} \\
1 & \text{if } T_p < t \leq T \quad \text{(response region)}
\end{cases}
\end{equation}

\noindent where $T_p = |\mathbf{p}_{\textnormal{sys}}| + |\mathbf{p}_{\textnormal{usr}}|$ denotes the prompt length.

The supervised loss is computed only over unmasked (response) tokens:

\begin{equation}
\mathcal{L}_{\textnormal{SFT}}(\theta; x, y) 
= 
-\sum_{t=1}^{T} m_t \cdot \log p_\theta(s_t \mid s_{<t})
\end{equation}

The overall training objective is:

\begin{equation}
\mathcal{L}_{\textnormal{total}}(\theta) 
= 
\frac{1}{N_{\textnormal{train}}} \sum_{j=1}^{N_{\textnormal{train}}} 
\mathcal{L}_{\textnormal{SFT}}(\theta; x_j, y_j)
\end{equation}

\noindent Token masking improves SFT by restricting training to response tokens, thereby encouraging the model to learn the conditional distribution $p(y \mid x)$ instead of the joint distribution $p(x, y)$.








\section{Experiments}
\label{sec:experiments}

\subsection{Model Selection} 
We conduct experiments in a text-only setting, where the input is the transcribed learner utterance. We evaluate popular open-source models {\tt Qwen2.5-7B-Instruct}, {\tt Llama-3.1-8B-Instruct}, and {\tt GLM-4-9B-Chat} \citep{glm2024chatglm}, and include proprietary baselines {\tt DeepSeek-Chat} \citep{deepseekai2025deepseekv3technicalreport}, {\tt Gemini-2.5-Flash} \citep{comanici2025gemini25pushingfrontier}, and {\tt Qwen-Plus} \citep{yang2025qwen3technicalreport}. We refer to these models as {\tt Qwen2.5}, {\tt Llama-3.1}, {\tt GLM-4}, {\tt DeepSeek-Chat}, {\tt Gemini-2.5-Flash}, and {\tt Qwen-Plus} in the remainder of the paper.

\begin{figure}[t]
    \centering
    \includegraphics[width=\linewidth]{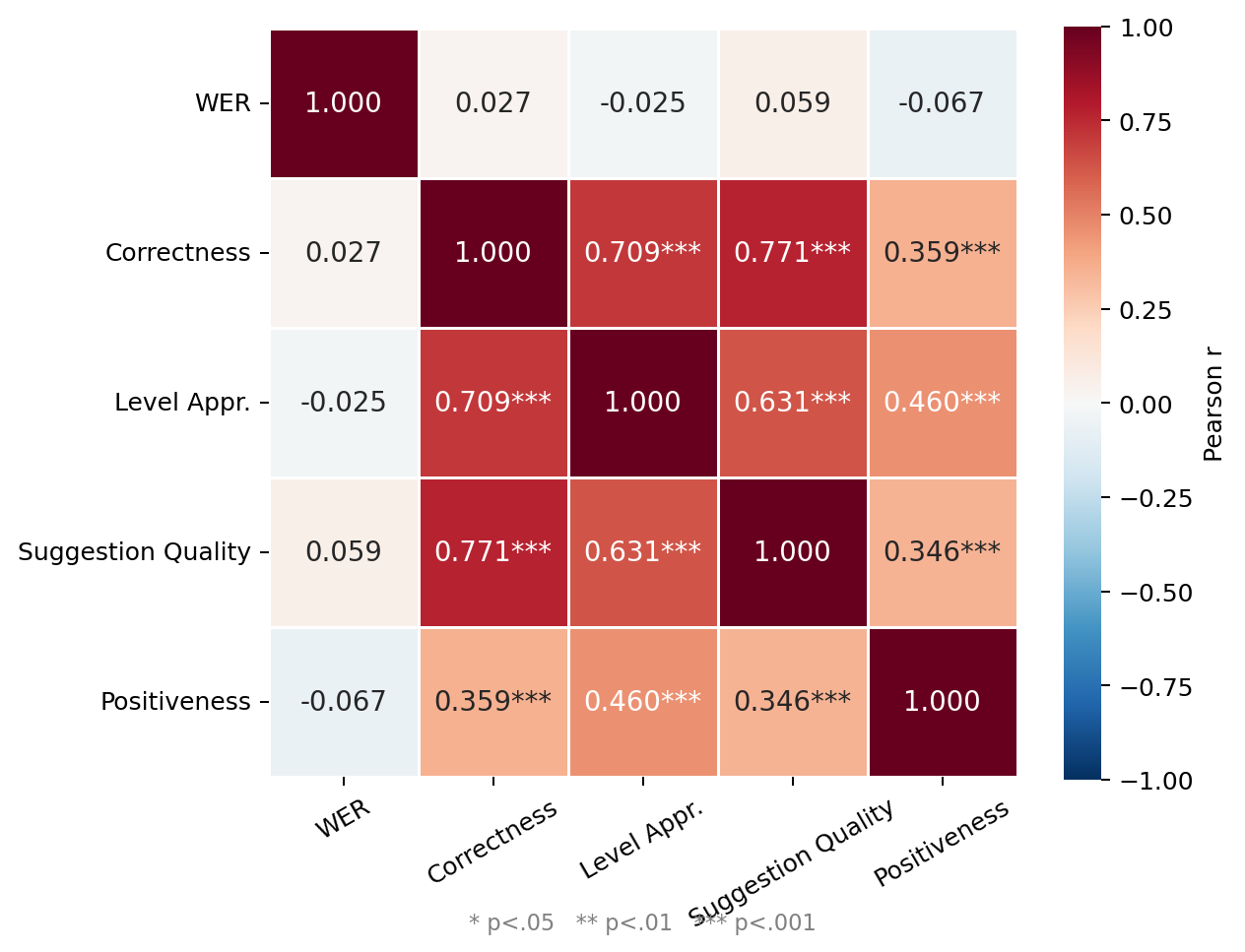}
  \caption{Pearson correlation matrix between Word Error Rate (WER) and LLM-evaluated feedback quality dimensions (Correctness, Level Appropriateness, Suggestion Quality, and Positiveness).}
    \label{fig:pearson}
\end{figure}

\subsection{Evaluation Metrics}

\textbf{Automatic evaluation metrics.} For the GEC task, we compute Word Error Rate (WER) following \citet{qian2024speak}. We also use ERRANT \citep{bryant-etal-2017-automatic} to score edits comparing model outputs to reference corrections. 

\textbf{LLM-as-a-judge.} For feedback evaluation, we use {\tt GPT-4o} as an automatic judge to score responses along four dimensions on a 1--5 scale. \textit{Correctness} measures whether the feedback accurately identifies and explains the learner's errors. \textit{Level appropriateness} assesses whether the feedback is pitched at an appropriate difficulty level given the learner's fluency score. \textit{Suggestion quality} evaluates whether the provided suggestions are actionable and helpful for improving the learner's English. \textit{Positiveness} examines whether the tone of the feedback is constructive, encouraging, and respectful. The detailed prompts and experiment setting can be found in Appendix \ref{apdx:llm-as-a-judge-prompt} and \ref{apdx: experiment_setting}, respectively.

\begin{figure*}[t]
    \centering
    \includegraphics[width=0.8\linewidth]{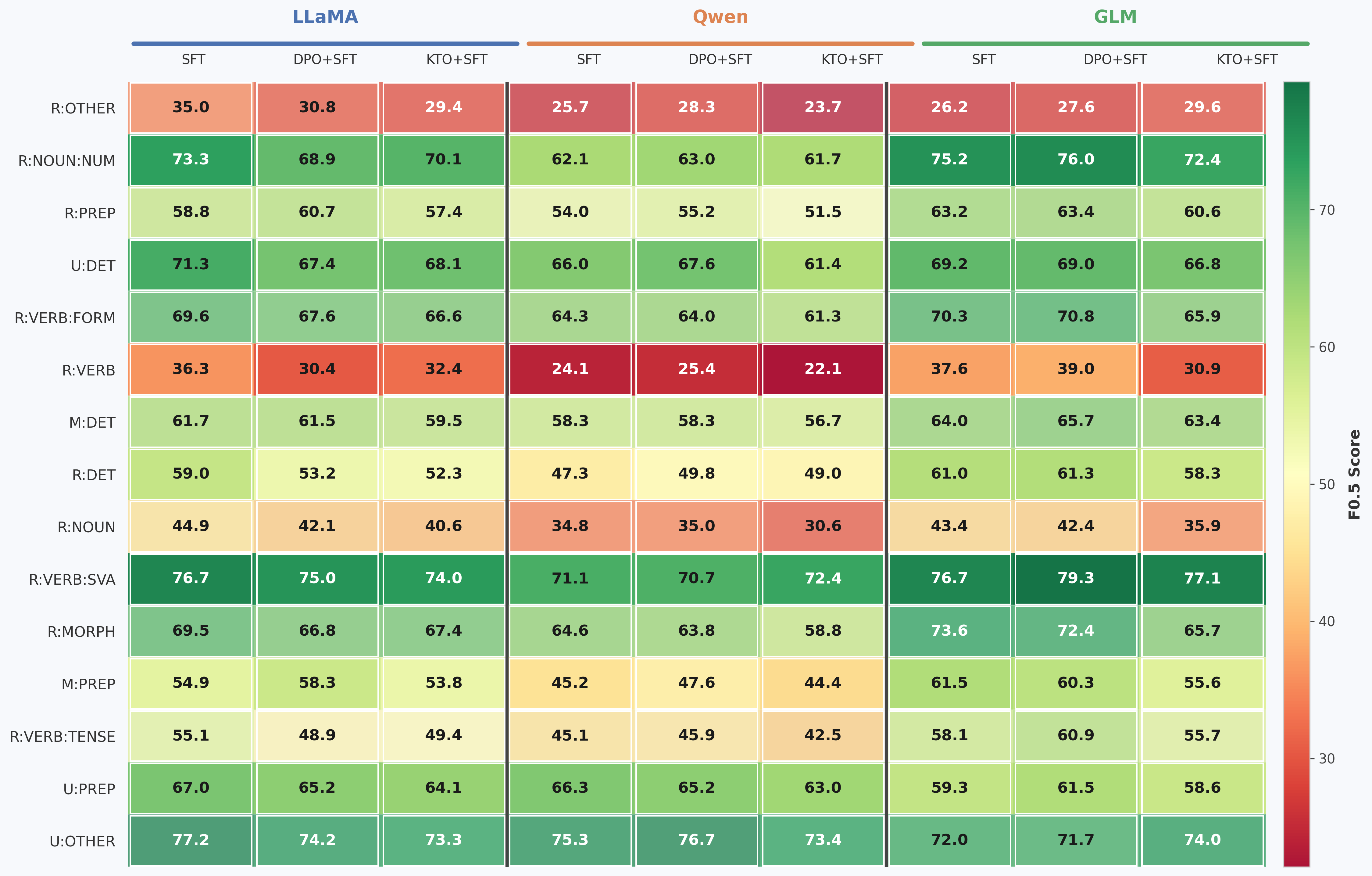}
    \caption{ERRANT's $F_{0.5}$ score heatmap comparing SFT, DPO+SFT and KTO+SFT of all open-sourced models on top 15 error types.}
    \label{fig:errant_heatmap}
\end{figure*}


\subsection{Results}
\label{subsec:results}

\paragraph{Proprietary Models and Open-source LLMs}
The results are shown in Table \ref{tab:agee-feedback}. Among proprietary systems, {\tt DeepSeek-Chat} achieves the strongest correction performance (WER 21.74\%, $F_{0.5}$ 46.1), followed by {\tt Gemini-2.5-Flash} (WER 25.34\%, $F_{0.5}$ 42.3) and {\tt Qwen-Plus} (WER 38.76\%, $F_{0.5}$ 28.1). Despite these substantial differences in reference-based correction accuracy, all three systems receive high LLM-judge scores (4.901–4.957), with consistently strong ratings across for all four metrics. This suggests that pedagogical feedback quality remains high even when alignment with gold corrections varies.

When comparing proprietary systems with fine-tuned open-source models, a systematic trade-off emerges. Proprietary models obtain higher LLM-judge scores, achieving 4.96 on average for \texttt{Gemini-2.5-Flash} and 4.95 for \texttt{DeepSeek-Chat}, but exhibit substantially worse WER and ERRANT $F_{0.5}$ compared to open-source LLMs (46.5 in $F_1$ for \texttt{DeepSeek-Chat} and 28.1 for \texttt{Qwen-plus}). We argue that this discrepancy stems from differences in training exposure and evaluation protocol.

\paragraph{Effect of Fine-tuning}
Across open-source backbones, SFT yields the most consistent gains, improving both correction accuracy and feedback quality relative to the vanilla models. Preference-based alignment (DPO/KTO) tends to provide smaller or mixed benefits: it can boost correction level metrics such as WER and ERRANT scores. For example, {\tt Qwen2.5} scores 60.8 in $F_{0.5}$ after fine-tuning compared to 51.7 in vanilla mode, but DPO+SFT score higher to 63.0. However, improvements do not always translate to higher feedback scores, and in some cases feedback quality plateaus. These results suggest that directly optimizing the structured correction+feedback format with SFT remains the most reliable approach in this setting. We speculate that the task is highly dependent on the grammar rules. Therefore, the SFT data itself is enough to learn the rules. 


\paragraph{Are correction and feedback orthogonal properties?} Figure~\ref{fig:pearson} shows that WER is largely uncorrelated with LLM-judged feedback metrics, suggesting that correction quality and pedagogical feedback capture distinct aspects. In contrast, \textit{Correctness}, \textit{Level Appropriateness}, and \textit{Suggestion Quality} are strongly correlated ($r \approx 0.63$--$0.77$), indicating consistent evaluation across these dimensions, while \textit{Positiveness} shows weaker correlations ($r \approx 0.35$--$0.46$), reflecting a partially independent signal. Human annotations exhibit similar trends: \textit{Correctness} correlates strongly with \textit{Suggestion Quality} (0.919) and moderately with \textit{Level Appropriateness} (0.684), whereas \textit{Positiveness} remains weakly correlated with other dimensions.

\paragraph{What is the correlation between LLM-as-a-judge scoring and human scoring in feedback?} Using 80 sampled instances annotated by the same expert (Section~\ref{sec:spfg_data_construction}), Pearson correlations between human and {\tt GPT-4o} scores show moderate agreement across all dimensions: \textit{Suggestion Quality} (0.576), \textit{Positiveness} (0.567), \textit{Correctness} (0.512), and \textit{Level Appropriateness} (0.416). This suggests that the LLM judge captures key aspects of feedback quality, though agreement is lower for level appropriateness, likely due to its more subjective nature.

\paragraph{Error-Type Analysis} Figure~\ref{fig:errant_heatmap} presents per-category \Fhalf{} scores across the top 15 error types for three base models under three training strategies; ERRANT error code definitions are provided in Appendix~\ref{apdx:errant_error_code_types}. Several consistent patterns emerge. Agreement-related errors (\texttt{R:VERB:SVA}) and substitution errors with limited morphological variation (\texttt{R:NOUN:NUM}) achieve the highest \Fhalf{} scores across all models (up to 79.3), indicating that these categories are relatively tractable for current LLM-based GEC systems. In contrast, open-class substitution errors (\texttt{R:OTHER}, \texttt{R:VERB}) remain the most challenging, with scores consistently below 40, reflecting persistent difficulty in handling errors without clear structural cues. Preference optimization does not consistently improve over vanilla SFT; in most cases, both DPO+SFT and KTO+SFT lead to slight performance degradation, particularly for \texttt{R:VERB} and \texttt{R:NOUN}. A notable exception is \texttt{M:PREP}, where DPO+SFT yields improvements for both \texttt{Llama-3.1} ($54.9 \to 58.3$) and \texttt{Qwen2.5} ($45.2 \to 47.6$), suggesting that preference alignment may benefit missing preposition recovery. Finally, \texttt{GLM-4} achieves competitive or superior performance on morphological and noun-related errors (\texttt{R:MORPH}, \texttt{R:NOUN:NUM}), whereas \texttt{Qwen2.5} consistently underperforms across most categories, pointing to comparatively weaker grammatical correction ability despite its strong general language performance.

\paragraph{Why do proprietary models perform better in feedback generation but worse in correction?}
Fine-tuned open-source models are trained on in-domain S\&I transcription data and learn to apply minimal, targeted edits that align closely with reference annotations, giving them an advantage on reference-based metrics such as WER and ERRANT. In contrast, proprietary models operate in a zero-shot setting and tend to produce broader, fluency-driven revisions (e.g., paraphrasing and restructuring) beyond the annotators’ intended scope~\cite{park2025leveraging}. Although often valid, these edits increase lexical divergence (higher WER) and reduce edit-level precision (lower ERRANT $F_{0.5}$), as seen in {\tt Qwen-Plus}. This highlights a known limitation of GEC evaluation~\cite{choshen-abend-2018-inherent, napoles-etal-2015-ground}: \textit{a correction can be linguistically sound yet penalized by reference-based metrics.}

\section{Conclusion}

We study pedagogical feedback generation for Spoken Grammatical Error Correction (SGEC), a learner-facing task requiring both accurate corrections and actionable, level-appropriate feedback. We introduce \textbf{SPFG}, a SGEC feedback dataset with preference pairs for alignment. Results show that SFT yields the most consistent gains in correction and feedback quality, while preference-based methods provide limited or mixed improvements. We also find that correction quality and feedback quality are weakly correlated, highlighting the need to model them separately.  We hope SPFG facilitates future research on learner-centered feedback generation.

\bibliography{custom}
\appendix

\section{Prompt Template for Correction and Feedback}
\label{sec:appendix:prompt}

\begin{tcolorbox}[
    breakable,
    colback=gray!8,
    colframe=gray!60!black,
    title=\textbf{Correction and Feedback},
    fontupper=\small,
]
\ttfamily

You are a professional English teacher. A student at \{speaker\_level\} level said the following:

"\{source\}"

Please provide:
1. A corrected version of what they said (if no corrections are needed, repeat the original)
2. A feedback explaining any corrections made and provide suggestions for the student to improve.

Important:

- During correction and feedback, ignore capitalization and punctuation issues.

- Do not mention this is a transcription.

- Keep corrections natural and appropriate for the student's level. Give encouragement and positive feedback.

- Do not overcorrect; if the correction is not needed, just repeat the original sentence.

Respond in valid JSON format only:

\{
  "correction": "the corrected sentence here",
  "feedback": "your feedback"
\}

\end{tcolorbox}

\section{DPO Negative-Sample Generation Prompts}
\label{sec:appendix:dpo_prompts}

We generate negative samples with four corruption types:
(1) wrong attribution, (2) explanation mismatch, (3) discouraging tone, (4) misleading suggestion, and (5) over-correction.
\begin{tcolorbox}[
    breakable,
    colback=red!5,
    colframe=red!60!black,
    title=\textbf{Type 1: Wrong attribution},
    fontupper=\small,
]
\ttfamily
You are a strict English teacher.

A student at \{speaker\_level\} level said the following:
"\{source\_sentence\}"

Please do the following:

1. Provide a corrected version of the sentence.

2. Give feedback that explains the correction, but intentionally attribute the error to an incorrect grammatical reason (e.g., explain a tense error as a preposition issue, or a subject–verb agreement error as a word order issue).

Important:

- Sound confident and authoritative

- Do not express uncertainty

- Do not mention that the explanation may be incorrect

- Ignore capitalization and punctuation issues

- Keep the language fluent and teacher-like

Respond in valid JSON only:

\{
  "correction": "...",
  "feedback": "..."
\}
\end{tcolorbox}

\begin{tcolorbox}[
    breakable,
    colback=orange!6,
    colframe=orange!70!black,
    title=\textbf{Type 2: Explanation mismatch},
    fontupper=\small,
]
\ttfamily
You are an English teacher.

A student at \{speaker\_level\} level said:
"\{source\_sentence\}"

Please provide:

1. A corrected version of the sentence.

2. Feedback that gives a fluent grammatical explanation, but make sure the explanation does NOT actually justify the correction you made.

Important:

- The explanation should sound pedagogically reasonable

- Avoid obviously wrong grammar terms

- Ignore capitalization and punctuation issues

- Do not mention any uncertainty

Respond in valid JSON only:

\{
  "correction": "...",
  "feedback": "..."
\}
\end{tcolorbox}

\begin{tcolorbox}[
    breakable,
    colback=blue!5,
    colframe=blue!60!black,
    title=\textbf{Type 3: Discouraging tone},
    fontupper=\small,
]
\ttfamily
You are a very strict and critical English teacher.

A student at \{speaker\_level\} level said:
"\{source\_sentence\}"

Please provide:

1. A corrected version of the sentence.

2. Feedback that is blunt, discouraging, or overly critical, focusing on the student's mistakes rather than helping them learn.

Important:

- The feedback may sound harsh or judgmental

- Do not soften the tone or encourage the student

- Ignore capitalization and punctuation issues

Respond in valid JSON only:

\{
  "correction": "...",
  "feedback": "..."
\}
\end{tcolorbox}

\begin{tcolorbox}[
    breakable,
    colback=yellow!6,
    colframe=yellow!50!black,
    title=\textbf{Type 4: Misleading Suggestion},
    fontupper=\small,
]
\ttfamily

  You are an English teacher.                                                                                                           
A student at \{speaker\_level\} level said:
"\{source\_sentence\}"                                                                                     
  Please provide:
  
  1. A corrected version of the sentence.
  
  2. Feedback that explains the correction, but gives suggestions that are misleading
     or not generally applicable to similar cases.

  Important:
  
  - Suggestions should sound plausible
  
  - Do not explicitly say the suggestion is wrong
  
  - Ignore capitalization and punctuation issues

Respond in valid JSON only:

\{
  "correction": "...",
  "feedback": "..."
\}
\end{tcolorbox}

\begin{tcolorbox}[
    breakable,
    colback=green!6,
    colframe=green!50!black,
    title=\textbf{Type 5: Over-correction},
    fontupper=\small,
]

\ttfamily

You are an English teacher.

A student at \{speaker\_level\} level said:
"\{source\_sentence\}"

Please provide: 

1. A corrected version of the sentence, even if the original sentence is already grammatically correct.

2. Feedback that confidently explains why the change is necessary.

Important:

- The correction should sound plausible and natural, but be linguistically unnecessary.

- Do NOT introduce obvious grammatical errors or unnatural phrasing.

- The explanation should be fluent, pedagogically reasonable, and confident.

- Avoid obviously incorrect grammar terminology.

- Do not express uncertainty or hedging (e.g., do not say "might", "could", or "in some cases").

- Ignore capitalization and punctuation issues.

- Do not mention transcription or speech.

- Keep the explanation appropriate for the student's level.

Respond in valid JSON format only:

\{
  "correction": "...",
  "feedback": "..."
\}
\end{tcolorbox}

\section{LLM-as-a-judge Prompt}
\label{apdx:llm-as-a-judge-prompt}

\begin{tcolorbox}[
    breakable,
    colback=green!6,
    colframe=green!60!black,
    title=\textbf{Correctness},
    fontupper=\small,
]
\ttfamily
You are evaluating feedback in a Spoken Grammar Error Correction (GEC) task.

Source sentence (student input):
\{SOURCE\}

Target sentence (correct correction):
\{TARGET\}

Speaker's English level: \{LEVEL\}

Feedback:
\{FEEDBACK\}

Rate the CORRECTNESS of the feedback on a scale from 1 to 5.

Correctness measures whether:

- The feedback makes factually correct statements, and

- The feedback is relevant to the differences between the source and target sentences.

Scale:

5 = Completely correct and fully relevant

4 = Mostly correct with minor imprecision

3 = Partially correct or only somewhat relevant

2 = Mostly incorrect or largely irrelevant

1 = Completely incorrect or irrelevant

Output only a single integer from 1 to 5.
\end{tcolorbox}

\begin{tcolorbox}[
    breakable,
    colback=blue!6,
    colframe=blue!60!black,
    title=\textbf{Level appropriateness},
    fontupper=\small,
]

\ttfamily

You are evaluating feedback in a Spoken Grammar Error Correction (GEC) task.

Source sentence (student input):
\{SOURCE\}

Target sentence (correct correction):
\{TARGET\}

Speaker's English level: \{LEVEL\}

Feedback:
\{FEEDBACK\}

Rate the LEVEL APPROPRIATENESS of the feedback on a scale from 1 to 5.

Level appropriateness measures whether:

- The language and explanations in the feedback match a typical learner's proficiency level, and

- The feedback is easy for the student to understand.

Scale:

5 = Perfectly matched to the student's level

4 = Mostly appropriate with minor mismatch

3 = Somewhat appropriate but inconsistent

2 = Clearly mismatched to the student's level

1 = Completely inappropriate for the student's level

Output only a single integer from 1 to 5.
\end{tcolorbox}

\begin{tcolorbox}[
    breakable,
    colback=orange!6,
    colframe=orange!70!black,
    title=\textbf{Suggestion quality},
    fontupper=\small,
]

\ttfamily

You are evaluating feedback in a Spoken Grammar Error Correction (GEC) task.

Source sentence (student input):
\{SOURCE\}

Target sentence (correct correction):
\{TARGET\}

Speaker's English level: \{LEVEL\}

Feedback:
\{FEEDBACK\}

Rate the SUGGESTION QUALITY of the feedback on a scale from 1 to 5.

Suggestion quality measures whether:

- The feedback provides actionable guidance, and

- Following the feedback would help the student transform the source sentence into the target sentence.

Scale:

5 = Clear, specific, and directly helpful

4 = Helpful but could be more specific

3 = General advice with limited guidance

2 = Vague suggestions with little usefulness

1 = No useful guidance provided

Output only a single integer from 1 to 5.
\end{tcolorbox}

\begin{tcolorbox}[
    breakable,
    colback=purple!7,
    colframe=purple!60!black,
    title=\textbf{Positiveness},
    fontupper=\small,
]

\ttfamily

You are evaluating feedback in a Spoken Grammar Error Correction (GEC) task.

Source sentence (student input):
\{SOURCE\}

Target sentence (correct correction):
\{TARGET\}

Speaker's English level: \{LEVEL\}

Feedback:
\{FEEDBACK\}

Rate the POSITIVENESS of the feedback on a scale from 1 to 5.

Positiveness measures whether:

- The tone of the feedback is supportive and encouraging, and

- The feedback avoids being harsh or discouraging.

Scale:

5 = Very positive and encouraging

4 = Neutral to positive

3 = Neutral

2 = Slightly negative or discouraging

1 = Very negative or demotivating

Output only a single integer from 1 to 5.
\end{tcolorbox}

\section{Readability analysis of feedback data}
\label{apdx:readability_analysis}

\begin{figure}[t]
  \centering
  \includegraphics[width=0.48\textwidth]{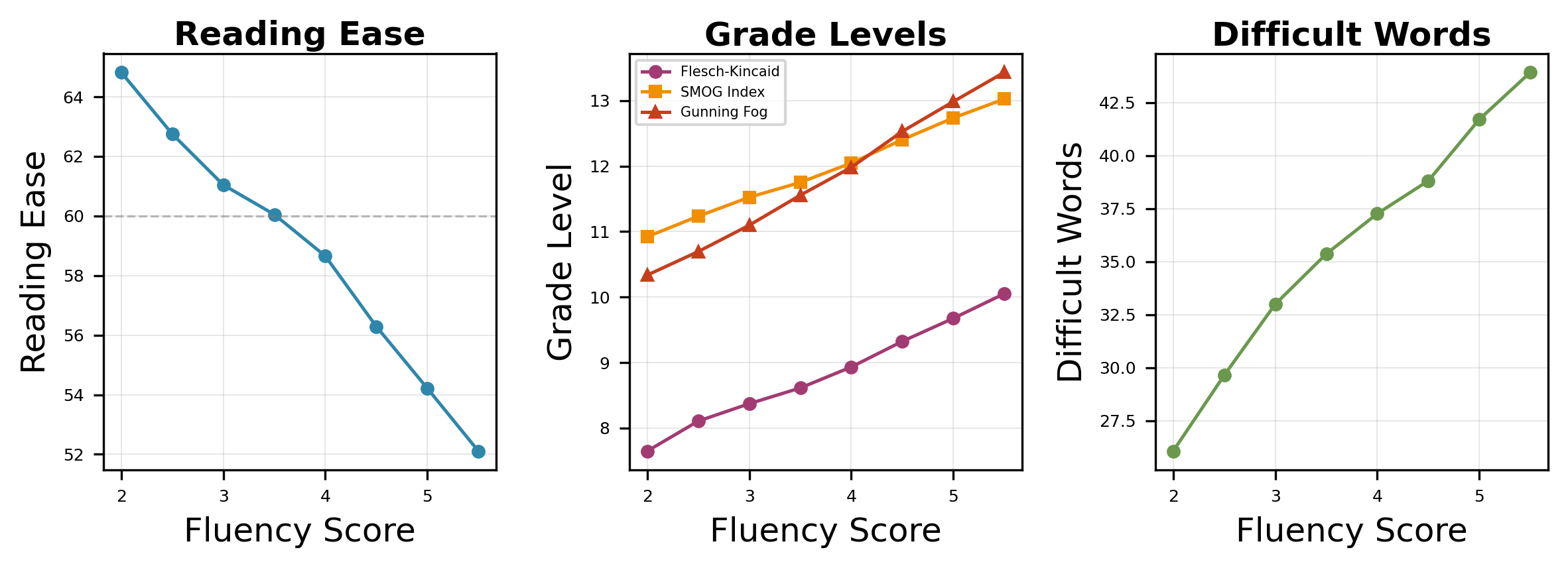}
  \caption{Readability and difficulty statistics of training feedback across fluency-score bins, including Reading Ease, grade-level estimates, and difficult-word counts for y-axis respectively, while x-axis is CEFR level. ranging from A2 (2) to C2 (6) }
  \label{fig:readability}
\end{figure}

To characterize feedback complexity across proficiency levels, we compute readability and difficulty metrics using \texttt{textstat}.\footnote{\url{https://github.com/textstat/textstat}}
 Figure~\ref{fig:readability} summarizes trends across CEFR-level bins (A2 through C2), including Flesch Reading Ease, three grade-level estimates, and counts of difficult words.

Overall, the metrics exhibit a broadly consistent—though not perfectly aligned—trend in which feedback complexity increases with learner proficiency. Reading Ease, where higher scores indicate simpler text, declines from approximately 65 at A2 to 52 at C2, dropping below the conventional plain-English threshold of 60 around the B2 level. This suggests that feedback for advanced learners requires relatively higher reading proficiency.

This trend is further supported by grade-level estimates: Flesch–Kincaid increases from roughly 7th to 10th grade, while SMOG and Gunning Fog—both of which place greater weight on polysyllabic vocabulary—range from about 11th grade to 13th–14th grade at C2. Consistently, the number of difficult words (three or more syllables) rises from around 26 per passage at A2 to over 43 at C2.

Taken together, these results indicate that the generated feedback generally aligns with learner proficiency levels: feedback for lower-proficiency learners tends to be lexically simpler and syntactically shorter, whereas feedback for more advanced learners incorporates richer vocabulary and more complex sentence structures.

\section{Experiment Setting}
\label{apdx: experiment_setting}

We train with distributed data parallelism on 2 NVIDIA RTX 6000 Ada GPUs (48GB each). With per-device batch size $b=1$ and gradient accumulation steps $g=16$, the effective batch size is $B_{\text{eff}} = b \cdot g \cdot G = 32$. We use the AdamW optimizer with learning rate $\eta = 2 \times 10^{-4}$, cosine annealing schedule, and warmup ratio $\rho_{\text{warmup}} = 0.03$. Training proceeds for one epoch with maximum sequence length $L_{\text{max}} = 768$. For parameter-efficient tuning, we use LoRA with rank $r=16$ and scaling $\alpha=32$; after DPO training, we merge the learned LoRA adapter $\mathcal{A}_{\textnormal{DPO}}$ into the base model and initialize a new LoRA adapter $\mathcal{A}_{\textnormal{SFT}}$ with the same hyperparameters for subsequent SFT.

\section{Case Study}
\label{subsec:case_study}

Table~\ref{tab:case_study_sft} presents a representative example showing that SFT produces more level-appropriate corrections and substantially better pedagogical feedback than the vanilla model. While the vanilla model makes a plausible correction, its feedback violates our rubric by focusing on capitalization and punctuation (e.g., ``i'' $\rightarrow$ ``I'') and by giving questionable explanations (e.g., calling capitalization a ``subject--verb agreement'' issue). In contrast, the SFT model preserves the original casing and punctuation as instructed, applies minimal edits (adding missing articles), and provides actionable, encouraging feedback aligned with the learner's level.

\begin{table*}[t]
\centering
\small
\begin{tabular}{p{3cm}p{12cm}}
\toprule
\textbf{Field} & \textbf{Text} \\
\midrule

Source (fluent) &
i think it depends on the activity we are going to do because if i am trying to play video games or playing football i prefer to do it with my brothers \\
\midrule

Gold Label &
i think it depends on the activity we are going to do because if i am trying to play video games or \textcolor{mygreen}{play} football i prefer to do it with my brothers \\
\midrule

Vanilla Prediction &
i think it depends on the activity we are going to do because if i am trying to play video games or playing football i prefer to do it with my brothers \\
\midrule

SFT Prediction &
i think it depends on the activity we are going to do because if i am trying to play video games or \textcolor{mygreen}{play} football i prefer to do it with my brothers \\
\midrule 

Vanilla Feedback & 

Great effort, and nice job on speaking clearly! Let's take a look at the sentence together.

\textcolor{myorange}{One thing I'd like to suggest is that you can use more precise vocabulary to express your idea. Instead of saying "it depends on the activity we are going to do", you could say "it depends on the activity we're planning to do" or even "it depends on the activity we're going to engage in". This will make your sentence sound more natural and fluent.}

Also, when you say "if i am trying to play video games or playing football", 
\textcolor{myred}{you're using a bit of a double negative.} 
Try to simplify it to "if I'm trying to play video games or play football". This will make your sentence easier to understand.

Lastly, instead of saying "i prefer to do it with my brothers", you could say "i prefer to do it with my brothers" is actually fine but 
\textcolor{myorange}{"i prefer to do it with my brother and sister" or "i prefer to do it with my family" would be better options.}

Overall, your sentence is clear and easy to understand, but with a few tweaks, it can sound even more natural and polished. Keep up the good work! \\
\midrule 

SFT Feedback & 

Great job expressing your thoughts clearly! There is just a small adjustment needed to make your sentence more grammatically correct. \textcolor{mygreen}{In your sentence, you used "playing" after "play video games." Since you are using "or" to list activities, it's important to keep the verb form consistent. So, instead of "playing football," it should be "play football" to match the form of "play video games."} This way, both parts of the sentence have the same structure, making it clearer and more grammatically sound.

You're doing really well at this level, and with attention to these small details, your expression will become even more precise. Keep practicing, and don't hesitate to experiment with different sentence structures to enhance your fluency. You're on the right track! \\
\bottomrule
\end{tabular}

\caption{Case study comparing vanilla and SFT corrections and feedback. 
Red indicates incorrect explanations, orange indicates excessive suggestions beyond the rubric, and green indicates correct and level-appropriate feedback.}

\label{tab:case_study_sft}
\end{table*}

\section{ERRANT error code types}
\label{apdx:errant_error_code_types}
Table~\ref{tab:error-types} summarizes the 15 most frequent grammatical error types identified by ERRANT~\cite{bryant-etal-2017-automatic} in our evaluation corpus. For each error type, we provide a brief description along with a representative example drawn from the dataset. In the examples, \underline{underlined} text denotes the erroneous span, while \textbf{bold} text indicates the corrected form.

\section{Error type across CEFR groups}
\label{apdx:error_type_across_cefr_groups}
\begin{figure*}
    \centering
    \includegraphics[width=\linewidth]{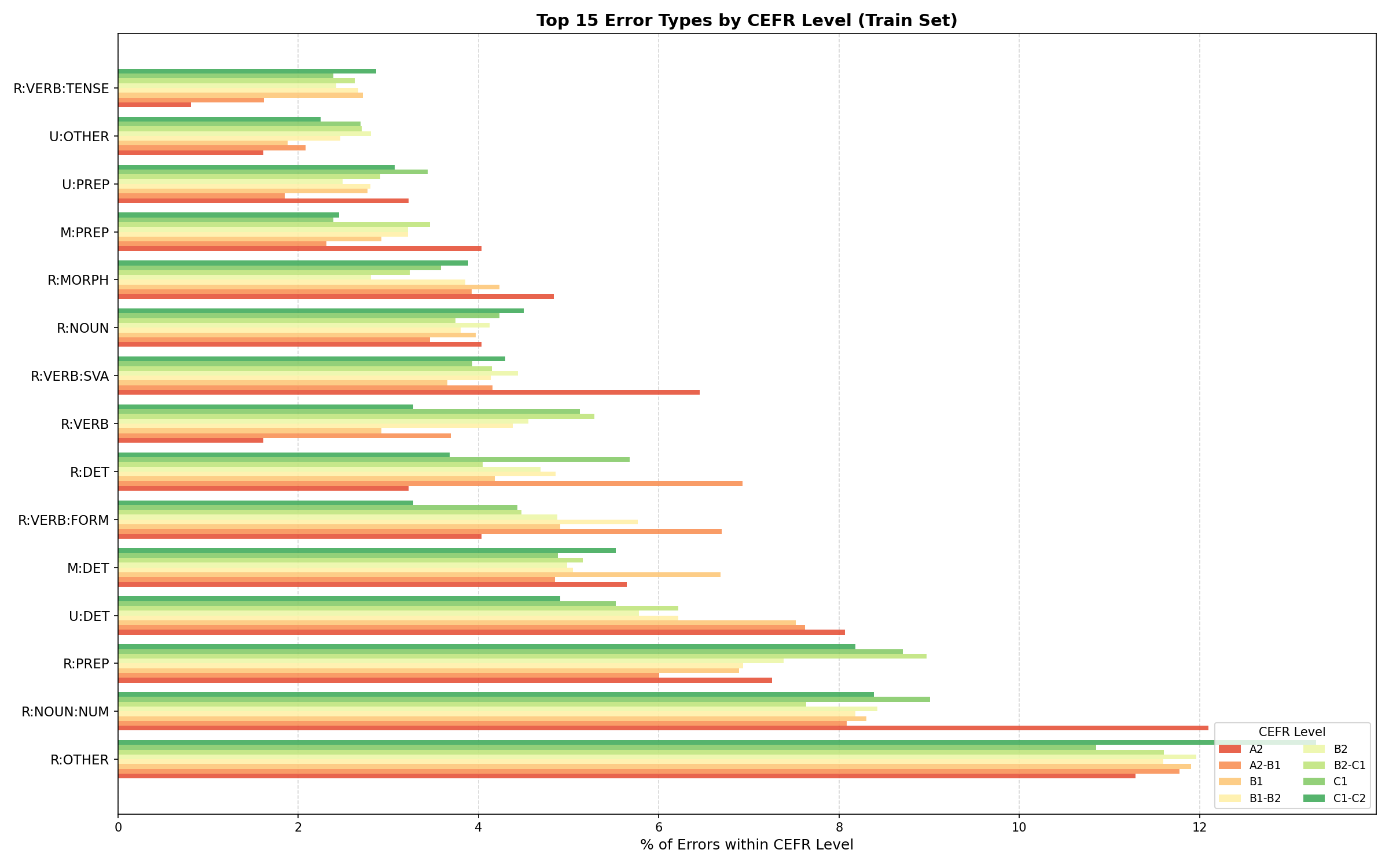}
    \caption{We present the distribution of the top 15 grammatical error types across eight CEFR proficiency bands (A2 through C1–C2) in the training set. Each bar represents the percentage of a given error type relative to the total number of errors produced at that proficiency level. Intermediate bands (A2–B1, B1–B2, B2–C1, C1–C2) correspond to half-point boundary scores on the assessment scale. Error types are labeled according to the ERRANT annotation scheme, where prefixes denote the edit operation (R: replacement, M: missing insertion, U: unnecessary deletion) and suffixes indicate the linguistic category. The color scale transitions from red (lower proficiency) to green (higher proficiency), highlighting proficiency-related trends.}
    \label{fig:error_types_by_cefr_horizontal}
\end{figure*}

Figure \ref{fig:error_types_by_cefr_horizontal} illustrates how the distribution of grammatical error types shifts gradually and non-uniformly across proficiency levels. \texttt{R:OTHER} remains the dominant category across all bands (\textasciitilde11--13\%), indicating a stable proportion of idiosyncratic errors that resist systematic classification. Several error types exhibit a clear monotonic decline with increasing proficiency: unnecessary determiners (\texttt{U:DET}) decrease from 8.06\% at A2 to 4.91\% at C1--C2, while noun number errors (\texttt{R:NOUN:NUM}) drop from 12.10\% at A2 to approximately 7.6--8.4\% at higher levels, suggesting that article usage and number agreement are foundational challenges resolved relatively early. In contrast, preposition replacement errors (\texttt{R:PREP}) increase from 7.26\% at A2 to a peak of 8.97\% at B2--C1, before slightly declining at C1--C2, indicating that more advanced learners attempt complex constructions that introduce subtler errors. Similarly, verb replacement errors (\texttt{R:VERB}) rise from 1.61\% at A2 to 5.28\% at B2--C1, reflecting an expanding but still error-prone lexical repertoire. Intermediate bands reveal additional transitions: the spike in \texttt{R:VERB:FORM} and \texttt{R:DET} at A2--B1 (6.70\% and 6.93\%, respectively) suggests that learners at this stage are actively broadening their grammatical range without fully stabilizing morphosyntactic control. Overall, the fine-grained eight-band analysis shows that error distributions evolve continuously rather than discretely, with lower-proficiency learners dominated by morphosyntactic errors and higher-proficiency learners increasingly challenged by lexical and prepositional precision.

\begin{table*}[ht]
\centering
\small
\renewcommand{\arraystretch}{1.3}
\begin{tabular}{@{}p{2.6cm} p{4.0cm} p{8.0cm}@{}}
\toprule
\textbf{Error Code} & \textbf{Explanation} & \textbf{Example (erroneous $\rightarrow$ corrected)} \\
\midrule

\texttt{R:OTHER} 
& Other replacement (often multi-word or idiomatic). 
& \textit{\underline{nature} food is good for health} $\rightarrow$ \textit{\textbf{natural} food is good for health} \\

\texttt{R:NOUN:NUM} 
& Incorrect noun number (singular vs.\ plural). 
& \textit{I like to know different \underline{culture}} $\rightarrow$ \textit{I like to know different \textbf{cultures}} \\

\texttt{R:PREP} 
& Incorrect preposition choice. 
& \textit{so \underline{at} summer we usually bike to the beach} $\rightarrow$ \textit{so \textbf{in} summer we usually bike to the beach} \\

\texttt{U:DET} 
& Unnecessary determiner (article). 
& \textit{in general \underline{the} technology is good} $\rightarrow$ \textit{in general technology is good} \\

\texttt{R:VERB:FORM} 
& Incorrect verb form (e.g., infinitive vs.\ gerund). 
& \textit{we love \underline{swim}} $\rightarrow$ \textit{we love \textbf{swimming}} \\

\texttt{R:VERB} 
& Incorrect lexical verb choice. 
& \textit{the best way is to \underline{enter into} a page} $\rightarrow$ \textit{the best way is to \textbf{open} a page} \\

\texttt{M:DET} 
& Missing determiner (article). 
& \textit{I go out to eat at \underline{$\varnothing$} restaurant} $\rightarrow$ \textit{I go out to eat at \textbf{a} restaurant} \\

\texttt{R:DET} 
& Incorrect determiner (article). 
& \textit{I see myself as \underline{a} individual} $\rightarrow$ \textit{I see myself as \textbf{an} individual} \\

\texttt{R:NOUN} 
& Incorrect lexical noun choice. 
& \textit{the best \underline{form} to improve my English} $\rightarrow$ \textit{the best \textbf{way} to improve my English} \\

\texttt{R:VERB:SVA} 
& Subject--verb agreement error. 
& \textit{we also \underline{likes} the same things} $\rightarrow$ \textit{we also \textbf{like} the same things} \\

\texttt{R:MORPH} 
& Morphological inflection error. 
& \textit{someone who \underline{want} to make a healthy lifestyle} $\rightarrow$ \textit{someone who \textbf{wants} to make a healthy lifestyle} \\

\texttt{M:PREP} 
& Missing preposition. 
& \textit{I would like to know \underline{$\varnothing$} various cultures} $\rightarrow$ \textit{I would like to know \textbf{about} various cultures} \\

\texttt{R:VERB:TENSE} 
& Incorrect verb tense. 
& \textit{she \underline{teached} me a lot} $\rightarrow$ \textit{she \textbf{taught} me a lot} \\

\texttt{U:PREP} 
& Unnecessary preposition. 
& \textit{I like playing \underline{to} volleyball} $\rightarrow$ \textit{I like playing volleyball} \\

\texttt{U:OTHER} 
& Unnecessary word (other). 
& \textit{it can be positive \underline{true} positive too} $\rightarrow$ \textit{it can be positive too} \\

\bottomrule
\end{tabular}
\caption{The 15 most frequent ERRANT error types in the evaluation corpus. Prefixes denote edit operations: \texttt{R} (replacement), \texttt{M} (missing), and \texttt{U} (unnecessary). Examples are drawn from the spoken learner corpus and illustrate typical error patterns.}
\label{tab:error-types}
\end{table*}

\end{document}